\documentclass[conference]{IEEEtran}
\IEEEoverridecommandlockouts
\usepackage{lipsum}
\usepackage{cite}
\usepackage{amsmath,amssymb,amsfonts}
\usepackage{balance}
\usepackage{algorithm}
\usepackage{subcaption}
\usepackage{relsize}
\usepackage{algpseudocode}
\usepackage{graphicx}
\usepackage{textcomp}
\usepackage{xcolor}
\def\BibTeX{{\rm B\kern-.05em{\sc i\kern-.025em b}\kern-.08em
    T\kern-.1667em\lower.7ex\hbox{E}\kern-.125emX}}

\renewcommand{\baselinestretch}{0.95}

\begin{document}

\title{HyperNQ: A Hypergraph Neural Network Decoder for Quantum LDPC Codes}


\author{
\IEEEauthorblockN{Ameya S. Bhave\IEEEauthorrefmark{1}, Navnil Choudhury\IEEEauthorrefmark{2}, Kanad Basu\IEEEauthorrefmark{2}}
\IEEEauthorblockA{\IEEEauthorrefmark{1}Department of Electrical and Computer Engineering, The University of Texas at Dallas, Richardson, TX, USA}
\IEEEauthorblockA{\IEEEauthorrefmark{2}Department of Electrical, Computer, and Systems Engineering, Rensselaer Polytechnic Institute, Troy, NY, USA}
}

\maketitle




\begin{abstract}

Quantum computing requires effective error correction strategies to mitigate noise and decoherence. Quantum Low-Density Parity-Check (QLDPC) codes have emerged as a promising solution for scalable Quantum Error Correction (QEC) applications by supporting constant-rate encoding and a sparse parity-check structure. However, decoding QLDPC codes via traditional approaches such as Belief Propagation (BP) suffers from poor convergence in the presence of short cycles. Machine learning techniques like Graph Neural Networks (GNNs) utilize learned message passing over their node features; however, they are restricted to pairwise interactions on Tanner graphs, which limits their ability to capture higher-order correlations. In this work, we propose HyperNQ, the first Hypergraph Neural Network (HGNN)--based QLDPC decoder that captures higher-order stabilizer constraints by utilizing hyperedges--thus enabling highly expressive and compact decoding. We use a two-stage message passing scheme and evaluate the decoder over the pseudo-threshold region. Below the pseudo-threshold mark, HyperNQ improves the Logical Error Rate (LER) up to 84\% over BP and 50\% over GNN-based strategies,  
demonstrating enhanced performance over the existing state-of-the-art decoders.

\end{abstract}


\section{Introduction} \label{sec:intro}

Quantum Error Correction (QEC) is essential for fault-tolerant quantum computing, protecting fragile quantum systems from gate faults, measurement errors, and decoherence~\cite{b1}.
QEC codes enable the detection and correction of errors by encoding the quantum information across multiple qubits, preserving the integrity of the encoded data. \textbf{Surface codes} are extensively studied among QEC codes~\cite{b2}. These codes benefit from planar layouts and low-weight checks, which made them practical in early hardware. However, they suffer from \textbf{low code rates}, incurring substantial \textbf{resource overhead}. To overcome this, asymptotically good quantum LDPC codes (\textit{e.g.} \textbf{Hypergraph Product (HGP) codes}) were developed to offer a scalable alternative~\cite{b4,b5}. Advances in chip design and qubit connectivity now position QLDPC codes as a more suitable choice at a hardware-level~\cite{b32}.

\begin{figure}[t]
    \centering
    \begin{subfigure}{0.49\columnwidth}
        \centering
        \includegraphics[width=\linewidth, height=4.5cm]{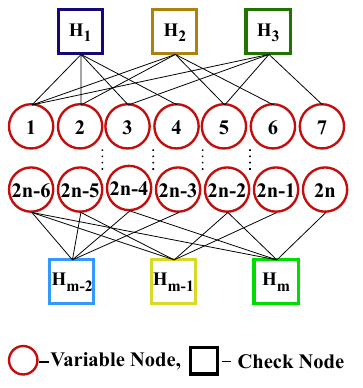}
        \caption{Tanner Graph}
        \label{fig:tanner_graph}
    \end{subfigure}
    \begin{subfigure}{0.49\columnwidth}
        \centering        \includegraphics[width=\linewidth, height=4.5cm]{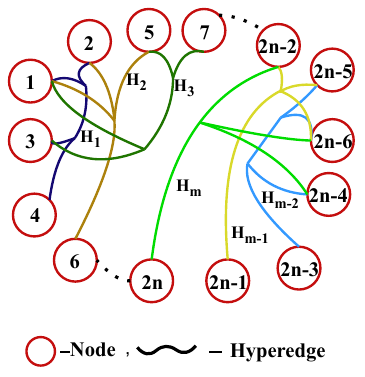}
        \caption{Hypergraph Representation}
        \label{fig:hypergraph}
    \end{subfigure}
    \caption{(a) represents the Tanner Graph and (b) is its Hypergraph Representation with weight (connected nodes) 4. Both graphs consist of $2n$ variable nodes (one per X- and Z-component of each qubit) and $m$ check nodes (stabilizers).} 
    \vspace{-6mm}
    \label{fig:intro_fig_1}
\end{figure}

QLDPC codes extend classical LDPC principles to quantum systems using sparse graph structures. They are commonly represented as \textbf{Tanner graphs}~\cite{b6,b8}. This is illustrated in \textbf{Fig.~\ref{fig:tanner_graph}}, where \textbf{circles} denote \textbf{$\textbf{1$\ldots$2n}$ variable nodes} (one per $X$- and $Z$-component of each qubit) and \textbf{squares} denote \textbf{H}$_{1}\ldots$\textbf{H}$_{m}$ check nodes (stabilizers), with edges indicating qubit stabilizer participation; further explained in Section~\ref{subsec:background2}. 
Classical and ML-based decoders use Tanner graphs to perform QEC. However, Tanner graphs are limited to modeling pairwise interactions, and hence these decoders fail to capture the multi-qubit stabilizer constraints inherent in QLDPC
code. The central challenge is to design a decoder that captures higher-order stabilizer constraints while maintaining linear scaling in block length. Explicit higher-order formulations, such as factor graphs or multi-edge expansions, lift the pairwise restriction, but they incur substantial computational cost and scale poorly~\cite{b14,b18}. Furthermore, BP-based decoding methods suffer from short cycles and poor convergence. Post-processing BP with Ordered Statistics Decoding (BP+OSD) partially mitigates these issues but introduces additional computational overhead~\cite{b8}. Machine learning (ML) decoders like 
Graph Neural Network (GNN)-based decoders further generalize BP by replacing fixed message passing rules with neural networks operating on Tanner graphs~\cite{b12,b15}. 
For any QEC code, decoder performance is inherently bounded by the \emph{pseudo-threshold}---the physical error rate \(p_f\) at which the logical error rate (LER) equals \(p_f\). Below this point, decoding yields net logical error suppression; above it, the encoded system can perform worse than leaving qubits uncoded.


In this paper, we introduce \textbf{HyperNQ}, a scalable and expressive decoding framework that, for the first time, applies a Hypergraph Neural Network (HGNN) architecture to QLDPC codes. Our study focuses on the \textbf{HGP code}, which is selected for its asymptotically favorable properties over surface codes, as further explained in Section~\ref{subsec:background2}.
Hypergraphs use hyperedges to connect multiple nodes (qubits), directly modeling the higher-order dependencies essential to quantum stabilizer codes~\cite{b21}. As illustrated in \textbf{Fig.~\ref{fig:hypergraph}}, hyperedges generalize conventional edges by removing the constraint of pairwise connectivity~\cite{b13}. HyperNQ uses this structure via a node$\to$hyperedge$\to$node message passing mechanism, enabling feature propagation, aggregation, and update across both nodes and hyperedges within a single layer. This design enables HyperNQ to represent multi-qubit stabilizer constraints between the nodes and hyperedges. For evaluation, we focus on the range spanning the region above and below the pseudo-threshold of our HyperNQ framework.

The key contributions of the paper are as follows:
\begin{itemize} 
\item \textbf{HGNN-based QLDPC Decoder:} We introduce HyperNQ, a novel QLDPC decoding framework that models multi-qubit stabilizers as hyperedges and uses an HGNN-based decoder to accurately capture higher-degree quantum parity constraints, enhancing decoding performance. 
\item \textbf{Two-Stage Message Passing (MP) Scheme:} 
We develop a node–hyperedge–node message passing mechanism for HyperNQ, incorporating attention and normalization, where the qubit (node) and stabilizer (hyperedge) features are aggregated and updated, enabling high expressivity.
\item \textbf{Evaluation:} 
Our experimental evaluations demonstrate that, below the pseudo-threshold, HyperNQ achieves an improved logical error rate of up to $84\%$ over classical (BP, BP+OSD) and ML-based (GNN) decoders.
\end{itemize}

\section{Background : Code design and decoding} \label{sec:background}

\subsection{Quantum Stabilizer Codes}


QLDPC decoding utilizes the stabilizer formalism, identifying errors via syndrome measurements from commuting Pauli operators~\cite{b6}. Stabilizer codes extend classical codes by defining commuting Pauli operators that preserve quantum states. Errors that anticommute with these operators yield measurable syndromes~\cite{b5}. 

\subsubsection{Pauli Group and Stabilizer Formalism} \label{subsubsec:pauligrp_stform}
Errors in quantum computation can be represented using elements of the \textit{Pauli group} (\( \mathcal{P}_n \)), which consists of tensor products of single-qubit Pauli operators \(\{I, X, Y, Z\}\) for $n$ qubits. A stabilizer code is defined by a set of \( m \) independent commuting Pauli operators \( \{S_1, S_2, ..., S_m\} \), forming an Abelian subgroup \( S \) of \( \mathcal{P}_n \). Since each stabilizer imposes a constraint, the logical subspace has a dimension of \( 2^k \), where \( k = n - m \), leading to an \( [[n, k, d]] \) quantum code of distance \(d\)~\cite{b8}.



\subsubsection{Error Syndromes and Error Detection}



In stabilizer codes, errors are detected by measuring the syndrome associated with each stabilizer generator. For an error operator \( E \in \mathcal{P}_n \), acting on a codeword \( |\psi\rangle \), the syndrome corresponding to stabilizer \( S_i \) is defined as \( s_i = 0 \), if \( E \) commutes with \( S_i \), and \( s_i = 1 \), if \( E \) anticommutes with \( S_i \). The full syndrome vector \( \mathbf{s} = (s_1, s_2, \ldots, s_m) \) provides a \textit{measurement outcome} that identifies the error without collapsing the quantum state, enabling correction via a decoding algorithm~\cite{b26}.

\subsubsection{Calderbank–Shor–Steane (CSS) Codes} \label{subsubsec:csscode}
CSS codes are critical in QLDPC code construction. They separate parity-check constraints into independent classical binary codes for \( X \) and \( Z \) errors~\cite{b28}. 
Given two classical codes \( C_X \) and \( C_Z \) satisfying \( C_Z^\perp \subseteq C_X \), the stabilizer matrix takes the form:
\vspace{-2mm}
\begin{equation}
    \textstyle
    H = \begin{bmatrix} H_X & 0 \\ 0 & H_Z \end{bmatrix}
    \label{css_stabilizer_matrix}
\end{equation}
where \( H_X \) and \( H_Z \) detect \( X \)- and \( Z \)-errors, respectively.
CSS-based QLDPC codes enable efficient iterative decoding while preserving stabilizer conditions~\cite{b19}.

\subsection{Quantum LDPC Codes, Representation and Construction} \label{subsec:background2}



Quantum Low-Density Parity-Check (QLDPC) codes are defined by sparse stabilizer matrices \(H\) similar to Eq.~(\ref{css_stabilizer_matrix}), where each row represents a stabilizer check affecting a small group of qubits~\cite{b5}. Compared to surface codes, asymptotically good QLDPC codes achieve higher code rates and improved distance scaling, making them promising for fault-tolerant quantum computing~\cite{b4}.

QLDPC codes are typically represented using Tanner graphs, as shown in Figure~\ref{fig:tanner_graph}, where \(2n\) variable nodes (qubits) and \(m\) check nodes (stabilizers) form a bipartite structure. The graph includes: \textbf{(1) Variable nodes}, corresponding to the columns of the parity-check matrix \(H\); \textbf{(2) Check nodes}, corresponding to its rows; and \textbf{(3) Edges}, representing non-zero entries in \(H\) that indicate which qubits participate in which stabilizers~\cite{b8}. Check nodes enforce parity constraints through their connectivity to variable nodes.

Among various QLDPC constructions, Hypergraph Product (HGP) codes offer an effective trade-off between decoding complexity, distance scaling, and fault tolerance~\cite{b4}. Constructed by taking the tensor product of two classical LDPC codes with parity-check matrices \(H_1\) and \(H_2\), the resulting QLDPC matrices are constructed as,
\(H_X = [H_1 \otimes I;\, I \otimes H_2^T]\) and \(H_Z = [I \otimes H_2;\, H_1^T \otimes I]\).
This structure preserves stabilizer conditions while maintaining sparsity, making HGP codes well-suited for scalable quantum error correction.


\begin{figure*}[htp]
    \centering
    \includegraphics[width=0.95\textwidth, height=7cm]
    {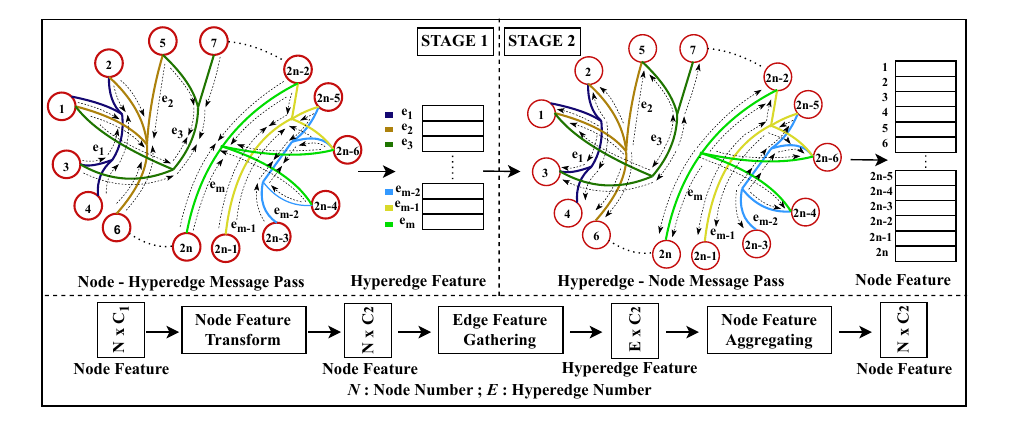}
    \vspace{-4mm}
    \caption{Illustration of the two-stage Node–Hyperedge–Node message passing mechanism. In the Node$\to$Hyperedge pass, hyperedge features are updated by aggregating with corresponding node features. In the Hyperedge$\to$Node pass, the updated hyperedge features are propagated back to update node features, ensuring effective modeling of multi-qubit parity constraints. 
    }
    \label{fig:methodology_fig_2}
    \vspace{-4mm}
\end{figure*}

\subsection{Decoding using BP, BP+OSD $\And$ ML-based Decoders} \label{subsubsec:bg_decd_bp_osd_ml}
Belief Propagation (BP) decodes QLDPC codes via an iterative message passing algorithm on Tanner graphs~\cite{b6, b26}. However, it encounters difficulties under high-noise conditions due to loops. The integration of BP with Ordered Statistics Decoding (BP+OSD) refines error estimates through soft-decision reordering, Gaussian elimination, and pattern estimation, thereby enhancing accuracy. However, this comes at the expense of increased computational complexity due to matrix inversion and combinatorial searches~\cite{b8, b20}. While ML-based methods, such as NBP and GNN-based decoders, improve convergence and generalization, their reliance on pairwise message passing restricts their ability to fully capture the complexity of stabilizer codes~\cite{b9, b10, b12, b15}. 

\subsection{Hypergraph Neural Network Overview}

HGNNs generalize the pairwise graph structures of the GNN into hypergraphs, where a single hyperedge connects multiple nodes to capture high-order relationships \cite{b22}. They are implemented using spatial-based (message passing) methods, where each layer consists of (1) \emph{node-to-hyperedge aggregation} and (2) \emph{hyperedge-to-node propagation} \cite{b21}. 
Prior research has demonstrated that HGNNs outperform traditional GNNs in capturing high-order dependencies, leading to better feature representation and model performance \cite{b23}.

The general message passing framework in hypergraphs utilizes higher-order relationships encoded in hyperedges. 
The message passing process involves two steps:
\begin{enumerate}
  \item \textbf{Node-to-Hyperedge Aggregation}: For each hyperedge \(b\), messages from neighboring nodes \(a \in N_v(b)\) are aggregated after transformation:
    \begin{equation}
    \textstyle
    m_b^t = \sum_{a \in N_v(b)} M_v^t (x_a^t); \; y_b^t = U_e^t (w_b, m_b^t)
    \label{eq:node-hyp_msg_ps}
    \end{equation}
    Here, \(x_a^t\) is the input feature of node \(a\) at layer \(t\), which is first transformed using a node-level message function \(M_v^t(\cdot)\). The resulting messages are aggregated by summation to form \(m_b^t\), which is then updated into the hyperedge feature \(y_b^t\) via a learnable hyperedge update function \(U_e^t(\cdot)\), incorporating the hyperedge weight \(w_b\).
  \item \textbf{Hyperedge-to-Node Propagation}: Each node \(a\) aggregates messages from hyperedges \(b \in N_e(a)\): 
    \begin{equation}
    \textstyle
    m_a^{t+1} = \sum_{b \in N_e(a)} M_e^t (x_a^t, y_b^t); \; x_a^{t+1} = U_v^t(x_a^t, m_a^{t+1})
    \label{eq:hyp-node_msg_ps}
    \end{equation}
    In this stage, node \(a\) receives messages from incident hyperedges using a function \(M_e^t(\cdot)\) combining the node’s current state \(x_a^t\) and each hyperedge feature \(y_b^t\). These are aggregated into \(m_a^{t+1}\) and used in the vertex update function \(U_v^t(\cdot)\) to compute the new node feature \(x_a^{t+1}\).
\end{enumerate}
\section{Proposed HyperNQ Framework} \label{sec:implementation}
We propose HyperNQ framework, which essentially recasts a tanner graph as a hypergraph at the core. This structure captures multi-qubit interactions, particularly within the CSS code formalism, where each \([[n, k, d]]\) code
induces \(2n\) variable nodes representing both \(X\)- and \(Z\)-type components~\cite{b4,b5}. By modeling these many-to-many relationships in a single structure, hypergraphs transcend pairwise constraints. 
The proposed two-stage HyperNQ decoding framework is depicted in Figure~\ref{fig:methodology_fig_2}, and the functional components under both stages are presented in Figure~\ref{fig:algorithm_flow_fig_3} and described subsequently.

\subsection{Incidence-Based Hypergraph Construction} \label{subsec:inc_hypcons}
We represent qubits as \textit{nodes} $(1\ldots2n)$ and stabilizers as \textit{hyperedges} $(e_{1}\ldots e_{m})$, encoded in an incidence matrix \( \mathbf{H} \in \{0,1\}^{2n \times m} \). In the matrix, \( 2n \) nodes are the combined \(X\) and \(Z\) errors, which correspond to the error operator, and \( m = m_x + m_z \) represents the total number of stabilizers. Each entry \( \mathbf{H}(i, j) = 1 \) signifies that node \( i \) participates in stabilizer \( j \). Next, we convert \( \mathbf{H} \) into a hyperedge index format (listing nodes associated with each hyperedge) to facilitate efficient computational message passing, as detailed subsequently.  


\subsection{Per-Node and Per-Hyperedge Feature Encoding} \label{subsec:node-hyp_featenc}
We represent each node $N$ by a structured feature vector containing a \textit{binary index encoding}, a \textit{bit\_value} indicator reflecting initial channel or syndrome information, and optional reliability metrics. In each hyperedge we  store a syndrome value ($\texttt{syndrome}_b$) and an associated weight ($w_b = 1 + \texttt{syndrome}_b$). The dimensionality of the node features before their aggregation into hyperedges is denoted $C_1$, while the dimensionality after the node feature transform stage is $C_2$ as shown in Figure~\ref{fig:methodology_fig_2}. This structured encoding, combined with explicit stabilizer weighting, emphasizes high-impact checks during message passing and supports soft-decision metrics (\textit{e.g.}, log-likelihood ratios) to improve decoding accuracy.

\subsection{Proposed Two-Stage Message Passing} \label{subsec:twostagemsgpass}
We design a novel HGNN decoder for HyperNQ that utilizes our proposed two-stage message-passing scheme within each layer. We use \textbf{Algorithm~\ref{alg:message_passing}} as the driver that executes \textbf{Algorithm~\ref{alg:node2hyperedge}} and \textbf{\ref{alg:hyperedge2node}} which correspond to \textbf{Stages 1} and \textbf{2} from \textbf{Figure~\ref{fig:methodology_fig_2}}. The inputs to Algorithm~\ref{alg:message_passing} include the node and hyperedge features (\textit{X} and \textit{Y}, respectively), the hyperedge weight ($w$), attributes ($S$), and index format (\textit{E}) of the incidence matrix (\textit{H}). The outputs are the updated node and hyperedge features ($X'$ and $Y'$). 
\begin{algorithm}[htbp]
\footnotesize
\caption{Two-Stage Message Passing (HyperNQ Layer)}
\label{alg:message_passing}
\begin{minipage}{\linewidth}
\textbf{Input:} Node features $X\in\mathbb{R}^{2n\times d_x}$, Hyperedge features $Y\in\mathbb{R}^{m\times d_h}$, 
Hyperedge connectivity $E\in\mathbb{N}^{2\times E}$, Hyperedge weights $w\in\mathbb{R}^{m}$, 
Hyperedge attributes $S\in\mathbb{R}^{m}$ (e.g., syndrome values) \\
\textbf{Output:} Updated node features $X'$ and hyperedge features $Y'$ 
\begin{algorithmic}[1]
    \Function{TwoStageMessagePassing}{$X,Y,E,w,S$}
        \State $Y' \gets \text{Node2Hyperedge}(X, Y, E, w, S)$ \hfill \textbf{// Stage 1, Algorithm~\ref{alg:node2hyperedge}}
        \State $X' \gets \text{Hyperedge2Node}(X, Y', E, w, S)$ \hfill \textbf{// Stage 2, Algorithm~\ref{alg:hyperedge2node}}
        \State \Return $X', Y'$
    \EndFunction
\end{algorithmic}
\end{minipage}
\end{algorithm}

\subsubsection{\textbf{Stage 1 : Node $\to$ Hyperedge Message Pass}} \label{subsubsec:nd-hyp_ms_pass}
\begin{figure}[t]
    \centering
    \includegraphics[width=\linewidth, height=4cm]{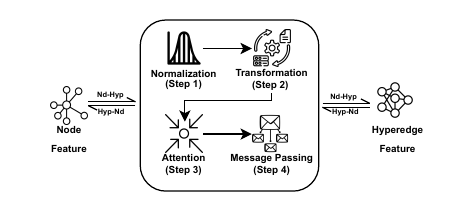}
    \vspace{-8mm}
    \caption{Block-level overview of Algorithm~\ref{alg:node2hyperedge} and \ref{alg:hyperedge2node} showing the 4-step process in both stages of the message-passing scheme.}
    \label{fig:algorithm_flow_fig_3}
    \vspace{-4mm}
\end{figure}

In Algorithm~\ref{alg:node2hyperedge}, we use a \textbf{four-step} process as shown in Figure~\ref{fig:algorithm_flow_fig_3}. We use this process over the node features to update the hyperedge feature representations. In \textbf{step 1} we compute a \textbf{normalization factor} \(B^{-1}(j)\) with Eq.~(\ref{eq:impl_normalization_node_hyp_impl_node-hyp_msg_ps}a). The denominator represents the degree of hyperedge \(j\) \textit{(lines 2--5)}.
\vspace{-1mm}
\begin{equation}
    a) \; \textstyle
    B^{-1}(j) \;=\; \frac{1}{\sum_{i:(i,j) \in E} 1}; \; b) \; \textstyle
    f_i \;=\; \sum_{i:(i,j) \in E} M_v(X_i)
    \label{eq:impl_normalization_node_hyp_impl_node-hyp_msg_ps}
\end{equation}
This operation mitigates bias toward hyperedges with higher connectivity. Without normalization, stabilizers involving many qubits would disproportionately influence message passing, inhibiting accurate error localization. In \textbf{step 2}, we \textbf{transform} the messages from nodes with Eq.~(\ref{eq:impl_normalization_node_hyp_impl_node-hyp_msg_ps}b),
where \(M_v(\cdot)\) transforms each node feature \(X_i\). This is displayed in Figure~\ref{fig:methodology_fig_2}, where node features are transformed from $(N\times C_{1})$ to $(N\times C_{2})$, where $N = 2n$ qubits (nodes) (\textit{lines 6--7}).
\begin{algorithm}[!htbp]
\footnotesize
\caption{Node $\to$ Hyperedge Message Pass}
\label{alg:node2hyperedge}
\begin{minipage}{\linewidth}
\textbf{Input:} Nd. features $X$, Hyp. features $Y$, connectivity $E$, wt. $w$, attr. $S$ \\
\textbf{Output:} Updated hyperedge features $Y'$
\begin{algorithmic}[1]
    \Function{Node2Hyperedge}{$X,Y,E,w,S$}
        \For{each hyperedge $j=1,\ldots,m$}
            \State $B(j) \gets \text{CountNeighbors}(E, j)$ 
            \State $B^{-1}(j) \gets \text{Inverse}(B[j])$ 
            \hfill \textbf{// 1: Normalization, Eq.~(\ref{eq:impl_normalization_node_hyp_impl_node-hyp_msg_ps}a)}
        \EndFor
        \For{each node $i$ s.t. $(i,j)\in E$}
            \State $f_i \gets \text{TransformNode}(X_i)$ 
            \hfill \textbf{// 2: Transformation, Eq.~(\ref{eq:impl_normalization_node_hyp_impl_node-hyp_msg_ps}b)}
            \If{$S$ exists}
                \State $\alpha_{ij} \gets  \text{Softmax(Score}(f_i, S_j))$ 
                \hfill \textbf{// 3: Attention,  Eq.~(\ref{eq:impl_attention_impl_node_hyp_update}a)}
                \State $f_i \gets \text{ApplyAttentionWeight}(f_i, \alpha_{ij})$ 
            \EndIf
            \State $f_i \gets
            \text{ApplyHyperedgeWeight}(f_i, w_{j})$
            \State $m_j \gets m_j + B^{-1}(j) \cdot f_i$
        \EndFor
        \For{each hyperedge $j$} 
            \State $Y'_j \gets \text{UpdateHyperedge}(Y_j,m_j)$ \hfill \textbf{// 4: Msg Passing, Eq.~(\ref{eq:impl_attention_impl_node_hyp_update}b) }
        \EndFor
        \State \Return $Y'$
    \EndFunction
\end{algorithmic}
\end{minipage}
\end{algorithm}
\vspace{-1.2ex}
Before finalizing the hyperedge representations, we add an \textbf{attention mechanism} in \textbf{step 3} to refine error sensitivity \cite{b21}. This is particularly important in quantum decoding, where certain syndromes indicate errors more reliably than others \cite{b19}. For each node-hyperedge pair the attention mechanism assigns importance scores, \(\alpha_{ij}\), as shown in Eq.~(\ref{eq:impl_attention_impl_node_hyp_update}a), where, $f_{i}$ represents the transformed node feature, while $S_{j}$, $S_{j'}$ denote the syndromes of hyperedge (stabilizer) $j$, $j'$. This enables the model to prioritize stabilizer--qubit relationships most indicative of error configurations \textit{(lines 8--11)}. 
\vspace{-1mm}
\begin{equation}
    a) \ \textstyle
    \alpha_{ij} = 
    \frac{\exp(\text{score}(f_i, S_j))}
         {\sum_{j':(i,j')\in E} \exp(\text{score}(f_{i}, S_{j'}))} \; ; \; b) \; \textstyle
    Y'_j \;=\; U_e(Y_j, m_j)
    \label{eq:impl_attention_impl_node_hyp_update}
\end{equation}
In the final step, \textbf{step 4}, we proceed to weight the final node$\to$hyperedge message by \(w_{j}\) and multiply it with the attention-scaled messages (\textit{lines 12--14}). This process improves error localization, guiding the decoder toward stabilizers most indicative of underlying qubit errors, which then aggregate into the hyperedge messages $m_j$. This \textbf{message-passing} is also depicted in Figure~\ref{fig:methodology_fig_2}, where the $j^{th}$ hyperedge with features $(E_{j}\times C_{2})$ are updated with the $i^{th}$ node having features $(N_{i}\times C_{2})$. Here, $E = m$, hyperedges are transformed by processing the aggregated messages as shown in Eq.~(\ref{eq:impl_attention_impl_node_hyp_update}b),
where \(U_e(\cdot)\) produces the updated hyperedge embedding (\(Y'_j\)), which is used as a more expressive stabilizer-level representation, integrating multi-qubit correlations crucial for accurate quantum decoding (\textit{lines 15--19}).
This entire aggregation process is mathematically expressed as Eq.~(\ref{eq:impl_nd_hyp_aggr}):
\vspace{-1mm}
\begin{equation}
    \textstyle
    H^e_j \;=\; 
    \sigma\!\Bigl(\,
      W^v 
      \sum_{\,i:(i,j)\in E} 
      \alpha_{ij}\,w_{j}\,X^v_i\,B^{-1}(j)
    \Bigr)
    \label{eq:impl_nd_hyp_aggr}
\end{equation}
where, \(H^e_j\) represents the updated feature of hyperedge \(j\). 
The transformation matrix (\(W^v\)) applies a learnable embedding to the node features (\(X^v_i\)), which are weighted by the attention coefficient (\(\alpha_{ij}\)) and scaled by the hyperedge weight (\(w_{j}\)). This formulation allows each stabilizer representation to capture intricate error syndromes, enhancing the decoder’s sensitivity to complex QLDPC error patterns. Finally, we apply a non-linear function \(\sigma(\cdot)\) (ReLU) to enhance the expressiveness of the hyperedge representation. 

\subsubsection{\textbf{Stage 2: Hyperedge $\to$ Node Message Pass}} \label{subsubsec:hyp-nd_ms_pass}

This stage focuses on reconstructing the qubit-level error information from stabilizer representations, as shown in Figure~\ref{fig:methodology_fig_2}. In QLDPC decoding, stabilizers provide essential multi-qubit syndrome data, but the ultimate objective is to identify error patterns at the qubit level.

The hyperedge$\to$node update in Algorithm~\ref{alg:hyperedge2node} ensures that high-order stabilizer insights are effectively propagated back to individual qubits.
\begin{algorithm}[!htbp]
\footnotesize
\caption{Hyperedge $\to$ Node Message Pass}
\label{alg:hyperedge2node}
\begin{minipage}{\linewidth}
\textbf{Input:} Nd. features $X$, Updated Hyp. features $Y'$, conn. $E$, wt $w$, attr. $S$ \\
\textbf{Output:} Updated node features $X'$
\begin{algorithmic}[1]
    \Function{Hyperedge2Node}{$X,Y',E,w,S$}
        \State FlipEdges($E$) 
        
        \For{each node $i=1,\ldots,2n$}
            \State $D(i) \gets \text{WeightedDegree}(E,i,w)$
            \State $D^{-1}(i) \gets \text{Inverse}(D[i])$ 
            \hfill \textbf{// 1: Normalization, Eq.~(\ref{eq:impl_normalization_hyp_node_impl_hyp-node_msg_ps}a)}
        \EndFor
    
        \For{each flipped edge $(j,i)\in E$}
            \State $g_j \gets \text{TransformHyperedge}(Y'_j)$ 
            \hfill \textbf{// 2: Transformation, Eq.~(\ref{eq:impl_normalization_hyp_node_impl_hyp-node_msg_ps}b)}
            \If{$S$ exists}
                \State $\beta_{ji} \gets \text{Softmax(Score}(g_j, S_j))$ 
                \hfill \textbf{// 3: Attention, Eq.~(\ref{eq:impl_attention_hyp_node_impl_hyp_node_update}a)}
                \State $g_j \gets \text{ApplyAttentionWeight}(g_j, \beta_{ji})$ 
            \EndIf
            \State $g_j \gets \text{ApplyHyperedgeWeight}(g_j, w_j)$
            \State $m_i \gets m_i + D^{-1}(i) \cdot g_j$
        \EndFor
    
        \For{each node $i$}
            \State $X'_i \gets \text{UpdateNode}(X_i,m_i)$ 
            \hfill \textbf{// 4: Msg Passing, Eq.~(\ref{eq:impl_attention_hyp_node_impl_hyp_node_update}b)}
        \EndFor
    
        \State \Return $X'$
    \EndFunction
\end{algorithmic}
\end{minipage}
\end{algorithm}
\vspace{-1.5ex}
We follow the same \textbf{four-step} process from Figure~\ref{fig:algorithm_flow_fig_3}, which performs the transformation of node features using the updated hyperedge features of stage 1.
In \textbf{step 1} we compute a \textbf{normalization factor} \(D^{-1}(i)\) for node \(i\), given by Eq.~(\ref{eq:impl_normalization_hyp_node_impl_hyp-node_msg_ps}a), where, the denominator is the weighted degree of hyperedge $j$ (\textit{lines 2--6}). This normalization balances contributions from multiple stabilizers. The nodes connected to many high-weight checks are prevented from disproportionately dominating the update. This ensures fair propagation of stabilizer information. 
\vspace{-1mm}
\begin{equation}
    a) \ \textstyle
    D^{-1}(i) \;=\; \frac{1}{\sum_{j:(j,i) \in E} w_{j}}; \; b) \ \textstyle
    g_j \;=\; \sum_{j:(j,i) \in E} M_e(Y'_j)
    \label{eq:impl_normalization_hyp_node_impl_hyp-node_msg_ps}
\end{equation}
In \textbf{step 2} we \textbf{transform} each hyperedge feature \(Y'_j\) via \(M_e(.)\), to incorporate it into the node representation (\textit{lines 7--8}) via Eq.~(\ref{eq:impl_normalization_hyp_node_impl_hyp-node_msg_ps}b).
In \textbf{step 3} we use an \textbf{attention mechanism} to assign importance scores to each hyperedge--node pair as shown in Eq.~(\ref{eq:impl_attention_hyp_node_impl_hyp_node_update}a), where, $g_{j}$ represents the transformed hyperedge (stabilizer) feature, while  $S_j$ and $S_{j'}$ represent the syndromes of hyperedge $j$ and $j'$, respectively \cite{b21}. 
This further enhances the model by tuning it to focus on the most significant hyperedge influences (\textit{lines 9--12}).
\vspace{-1mm}
\begin{equation}
    a) \ \textstyle
    \beta_{ji} 
    \;=\; 
    \frac{\exp(\text{score}(g_j, S_j))}
         {\sum_{j':(j',i) \in E} \exp(\text{score}(g_{j}, S_{j'}))}; \; b) \ \textstyle
    X'_i \;=\; U_v(X_i, m_i)
    \label{eq:impl_attention_hyp_node_impl_hyp_node_update}
\end{equation}
In the final step, \textbf{step 4}, we weight the aggregated message by \(w{_j}\) and incorporate the same into the node representation
(\textit{lines 13--15}). The \textbf{message-passing} step updates the node features via Eq.~(\ref{eq:impl_attention_hyp_node_impl_hyp_node_update}b),
where \(X'_i\) is the new node feature (\textit{lines 16--20}). As shown in Figure~\ref{fig:methodology_fig_2}, hyperedge features $(E_{j} \times C_{2})$ are aggregated into node features $(N_{i} \times C_{2})$, yielding refined qubit representations that combine local channel data with higher-order correlations derived from stabilizer syndromes. The transformation of hyperedge features back into node features follows Eq.~(\ref{eq:impl_hyp_nd_aggr}):
\vspace{-1mm}
\begin{equation}
    \textstyle
    H^v_i 
    \;=\;
    \sigma \!\Bigl(\,
      W^e 
      \sum_{\,j:(j,i)\in E} 
      \beta_{ji}\,w_{j}\,X^e_j\,D^{-1}(i)
    \Bigr)
    \label{eq:impl_hyp_nd_aggr}
\end{equation}
\vspace{-1mm}
where, \(H^v_i\) represents the updated feature of node \(i\). Here, the learnable transformation matrix
\(W^e\) applies an embedding to the aggregated hyperedge features \(X^e_j\). The coefficient
\(\beta_{ji}\) serves the same role as \(\alpha_{ab}\) but is indexed to reflect hyperedge-to-node
attention. The hyperedge weight \(w_{j}\) scales each hyperedge's contribution, and \(D^{-1}(i)\)
enforces balanced aggregation across nodes. Finally, the ReLU activation function \(\sigma(\cdot)\)
creates non-linearity in node representation. 
\section{Performance Evaluation} \label{sec:performance}
\subsection{Experimental Setup}
\subsubsection{\textbf{Benchmarks for Training}}
The training dataset comprises $2.5\times10^4$ syndrome-error pairs in the binary vector space $F^{2n}_2$, encapsulating both bit-flip (X) and phase-flip (Z) errors over $n$ qubits~\cite{b1}. Deterministic single-qubit and zero-error states ensure baseline coverage, supplemented by additional random errors sampled from an exponentially decaying distribution to realistically simulate quantum noise~\cite{b2}.

\subsubsection{\textbf{Benchmarks for Evaluation}}
For evaluation, we utilize a test dataset comprising $10^6$ syndrome-error pairs generated using an independent and identically distributed Pauli error model, enabling robust and scalable performance assessments across different physical error rates~\cite{b31}. To preserve evaluation integrity, the test dataset is independently sampled with a non-overlapping error distribution, ensuring zero intersection with the training set. The larger test set ensures high-confidence error rates, while training on $2.5\times10^4$ samples balances generalization and overfitting concerns. Each data point in the result is plotted with 95\% confidence intervals, obtained from repeated Monte Carlo trials, ensuring that the observed improvements are statistically significant.


\subsubsection{\textbf{Evaluation Metrics}}
We assess the decoder performance using Logical Error Rate (LER), the fraction of incorrect logical decodings. To quantify improvements, we compare state-of-the-art baseline decoders (\(\text{LER}_{\text{baseline}}\)) explained in Section~\ref{subsec:baseline_decoder_comp} with HyperNQ using:
\vspace{-1mm}
\[   
\text{LER improvement (\%)} = \frac{\text{LER}_{\text{baseline}} - \text{LER}_{\text{HyperNQ}}}{\text{LER}_{\text{baseline}}} \times 100 
   \label{eq:LER_improvement}
\]
Evaluating LER across varying physical error rates ($p_f$) reveals decoder robustness under realistic quantum noise.





\subsection{Baseline Decoding Approaches for Comparison} \label{subsec:baseline_decoder_comp}

To evaluate our HyperNQ framework, we compare its performance against the following state-of-the-art decoding approaches already described in Section~\ref{subsubsec:bg_decd_bp_osd_ml}.:

\begin{itemize} 
\item \textbf{\textit{Belief Propagation (BP)}}: A classical iterative decoder ~\cite{b8,b18}, adapted for quantum LDPC codes.

\item \textbf{\textit{BP + Ordered Statistics Decoding (BP+OSD)}}: Augments BP with post-processing~\cite{b8,b10}. Results are reported for both \textit{order 0} and \textit{order 4} to capture the performance--complexity trade-off.

\item \textbf{\textit{Graph Neural Network (GNN)-based decoder}}: an ML-based decoder on Tanner graphs that serves as the primary architectural baseline for comparison with our hypergraph-based approach~\cite{b12}.
\end{itemize}
\subsection{Decoder Performance and Comparative Analysis}

The proposed HGNN decoder in HyperNQ comprises a \textit{single} message passing layer as compared to the GNN decoder, which comprises \textit{six} layers \cite{b12}. We use the \textit{Binary Cross Entropy (BCE)} loss, optimized through the \textit{Adam optimizer} for both\cite{b12}. The model is evaluated using the following hyperparameters: a hidden dimension of \textbf{$\textit{128}$}, batch size of \textbf{$\textit{64}$}, learning rate of \textbf{$\textit{5} \times \textit{10}^{-\textit{5}}$}, and weight decay of \textbf{$\textit{5} \times \textit{10}^{-\textit{4}}$}. 

\begin{figure}[t]
    \centering
    \includegraphics[width=.9\columnwidth, keepaspectratio]
    {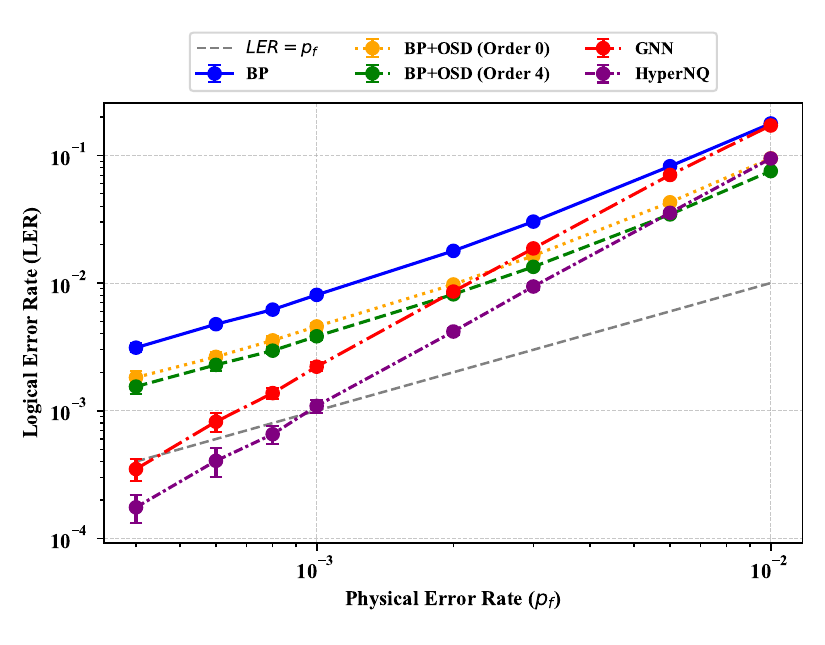}
    \vspace{-4mm}
    \caption{Performance comparison of our proposed HyperNQ versus other approaches across different physical error rates for decoding the HGP code, [[n, k]] = [[129, 28]]).}
    \label{fig:hgp_n_129_k_28_result}
    \vspace{-6mm}
\end{figure}

To evaluate the proposed decoder, we use a QLDPC code constructed via the hypergraph product (HGP) method with parameters $[[n, k]]=[[129,28]]$, based on component codes $H_1=[[7,4,3]]$ and $H_2=[[15,7,5]]$, as explained in Section~\ref{subsec:background2}. This construction offers a practical balance of length, code rate, and minimum distance for benchmarking~\cite{b8,b10,b12}. Figure~\ref{fig:hgp_n_129_k_28_result} presents the results, with Physical Error Rate ($p_f$) on the x-axis and Logical Error Rate (LER) on the y-axis.
We represent the pseudo-threshold boundary (LER = $p_f$) with a dashed gray line, and we mark the pseudo-threshold for the decoder where its curve intersects this line. Decoding is effective when LER $< p_f$. This indicates a net error suppression relative to the uncoded case. Decoders work to keep LER below their threshold, and a higher crossing indicates a larger operating window for practical QEC. By showing both the regions--above and below pseudo-threshold boundary--Figure~\ref{fig:hgp_n_129_k_28_result} captures the full decoder behavior, with the latter marking the practical region where HyperNQ proves effective. As seen in Figure~\ref{fig:hgp_n_129_k_28_result}, HyperNQ achieves net error suppression and hits the pseudo-threshold mark at $p_f = 0.001$, while GNN achieves it at about $p_f = 0.0005$. Classical decoders fail in the detection regime, remaining above the pseudo-threshold boundary and yielding logical error rates higher than the physical error rate. With 2$\times$ better pseudo-threshold over GNN, HyperNQ demonstrates substantial performance gains below $p_f = 0.001$. It achieves up to a \textbf{\textit{84}$\%$} lower LER compared to Belief Propagation (BP)~\cite{b8}, and outperforms on the BP+OSD baseline by \textbf{\textit{72}$\%$} (order 4) and \textbf{\textit{76}$\%$} (order 0) respectively~\cite{b8,b10}. 
Furthermore, HyperNQ achieves up to a \textbf{\textit{50}$\%$} reduction in LER relative to a state-of-the-art GNN decoder, underscoring its ability to capture complex, high-order correlations through hypergraph-based representations.

\subsection{Scalability and Computational Complexity}

HyperNQ scales linearly in blocklength inference. The hyperedges aggregate higher-order constraints in one hop, thus reducing per-layer propagation cost and layer count. In contrast, Tanner-graph GNNs also scale as $O(n)$ but require two pairwise hops,  (node$\to$edge) and (edge$\to$node) per layer; BP scales in $O(n)$ yet suffers from short-cycle effects while BP+OSD improves accuracy at exponential cost $O(k^{o})$ in the OSD order $o$ \cite{b12}. Thus, HyperNQ preserves linear scaling like BP/GNN while avoiding the $O(k^{o})$ blow-up and directly modeling multi-qubit constraints in one hop.
 Formally, for HyperNQ, let $H\in\{0,1\}^{2n\times m}$ denote the qubit–stabilizer incidence matrix for a CSS code
($2n$ variable nodes; $m$ checks).
Let $I=\mathrm{nnz}(H)$ be the total number of nonzero entries in $H$,  
i.e., the total qubit–stabilizer connections across both $X$ and $Z$ checks. Let $d$ be the hidden feature dimension.  
HyperNQ performs a complete \emph{node$\to$hyperedge$\to$node} message-passing cycle in one hop, 
where messages are exchanged only along these $I$ connections, giving a sparse aggregation 
cost $O(I\,d)$~\cite{b13}. After aggregation, each hyperedge and node applies a small dense transform (\textit{e.g.}, linear/MLP, incurring $d^2$ from $d\times d$ weight multiplication), costing $O(m\,d^2)$ and $O(2n\,d^2)$, respectively. The per-layer inference cost is:
\vspace{-1.3ex}
\[
T_{\text{HyperNQ}} = O(I\,d) + O\big((2n+m)\,d^2\big)
\]
For LDPC families with bounded check weight and bounded variable degree, we have
$I=\Theta(n)$ and $m=\Theta(n)$, yielding \emph{linear} scaling in blocklength for fixed $d$:
\vspace{-1.3ex}
\[
T_{\text{HyperNQ}} = \Theta(n\,d) + \Theta(n\,d^2)
\]
This efficiency is most impactful when stabilizers couple multiple qubits, where the hyperedge update offers the largest representational gain; in very low-weight layouts, the gap to pairwise schemes naturally narrows.

\section{Conclusion} 
\label{sec:conclusion}


In this work, we introduced HyperNQ, the first Hypergraph Neural Network (HGNN)–based decoder for Quantum LDPC codes, designed to model stabilizer constraints as hyperedges and capture higher-order multi-qubit correlations. 
The two-stage message passing architecture enables expressive and scalable decoding with linear complexity. 
When compared against state-of-the-art decoders —including BP, BP+OSD, and GNN-based models—HyperNQ achieves significantly lower Logical Error Rates (up to 84\% improvement over BP and 50\% over GNN-based decoders) and reduces computational overhead via shallower network depth under the pseudo-threshold regime. 
These results establish HyperNQ as a promising decoding framework for QLDPC codes, underscoring their utility in Quantum Error Correction. Future works can include exploring quantized inference on specialized accelerators for low-latency deployments and adaptive transfer learning across code families to extend generalization beyond a single topology.




\vspace{12mm}

\end{document}